# APPLYING DYNAMIC MODEL FOR MULTIPLE MANOEUVRING TARGET TRACKING USING PARTICLE FILTERING

Mohammad Javad Parseh[1] and Saeid Pashazadeh[2]

[1]M.Sc. Student of Artificial Intelligence
Faculty of Electrical and Computer Engineering, University of Tabriz, Tabriz, Iran
`m_paresh89@ms.tabrizu.ac.ir`
[2]Assistant Professor of Software Engineering
Faculty of Electrical and Computer Engineering, University of Tabriz, Tabriz, Iran
`pashazadeh@tabrizu.ac.ir`

*ABSTRACT*

*In this paper, we applied a dynamic model for manoeuvring targets in SIR particle filter algorithm for improving tracking accuracy of multiple manoeuvring targets. In our proposed approach, a color distribution model is used to detect changes of target's model. Our proposed approach controls deformation of target's model. If deformation of target's model is larger than a predetermined threshold, then the model will be updated. Global Nearest Neighbor (GNN) algorithm is used as data association algorithm. We named our proposed method as Deformation Detection Particle Filter (DDPF). DDPF approach is compared with basic SIR-PF algorithm on real airshow videos. Comparisons results show that, the basic SIR-PF algorithm is not able to track the manoeuvring targets when the rotation or scaling is occurred in target's model. However, DDPF approach updates target's model when the rotation or scaling is occurred. Thus, the proposed approach is able to track the manoeuvring targets more efficiently and accurately.*

## 1. INTRODUCTION

Multi target tracking is one of the important research areas in computer vision. Human tracking in monitoring applications and aircraft tracking in military applications are known as important applications areas of multi target tracking. Multi target tracking includes two main stages: 1) pose estimation and 2) data association. Kalman filtering and grid-based methods are the most important methods for pose estimation that are used in linear systems with Gaussian noises. For linear-Gaussian systems, Kalman filter is an analytical and optimal method. Particle filtering is a numerical and suboptimal method that is used for pose estimation. Unlike the Kalman filtering, Particle filtering method can be used in non-linear system with non-Gaussian noises. This is the

DOI:10.5121/ijitca.2012.2404        37

most important advantage of particle filtering over other methods. Estimation of manoeuvring target's state is one of the most important problems in pose estimation. Multiple Model (MM) approach is a common solution to resolve this problem [1-3]. Target's posteriori density function is represented by weighted sum of the output of several parallel filters in MM approach [4]. A real-time approach is presented in [15] for tracking a manoeuvring target by incorporating dynamic components in the target model. They employed particle filter based tracker. This tracker exploits a first order dynamic model and continuously performs adaptation of model noise for balancing uncertainty between the static and dynamic components of the state vector. The presented experimental evaluation shows that this approach is particularly effective in video sequences where appearance and motion have quick changes and there are partial or full target occlusions.

Data association algorithm associates each measurement to track. In cluttered environments, measurements may arise from detected targets or clutter. From the view of implementation, Global Nearest Neighbor (GNN) is the simplest method for data association [5]. Usually, this method is more efficient in clutter-free environments. One of the most common data association methods is Joint Probabilistic Data Association (JPDA) [6]. In this method, the probability of each measurement associated with each track is calculated. Multiple Hypothesis Tracking (MHT) is the most complex method for data association that is efficient for cluttered environments [7]. In MHT method, each hypothesis is a measurement-to-track association. Set of all hypothesizes is considered, and likelihood of each hypothesis is calculated. One of the best methods for obtaining best hypothesis is presented by Nummiaro et. al. [8]. In MHT method, the best hypothesis is chosen through the set of all hypothesizes by using a statistical approach, whereas in the JPDA method, an average of target state is calculated on all hypothesizes. This is the main difference between JPDA method and MHT method. In this paper, DDPF algorithm is proposed by improving the basic SIR algorithm for tracking the multiple manoeuvring targets using a dynamic model for targets.

Remaining sections of this paper is organised as follows: In section 2, particle filter method is introduced briefly. In section 3, basic SIR particle filter algorithm is introduced and in section 4, target representation model is explained. In section 5, a dynamic equation for manoeuvring target and in section 6, likelihood function is presented. In section 7, we briefly describe data association algorithm. In section 8, a new method is presented for deformation detection. In section 9, conditions of experiments is presented. In section 10, we compare results of our proposed DDPF method and basic SIR algorithm in multiple manoeuvring target tracking. Finally, in section 11, we'll discuss about and make conclusion from experimental results.

## 2. PARTICLE FILTER

Appropriate pose estimation method should be chosen according to tracking conditions. If the system is linear and system noise is Gaussian, Kalman filter is most appropriate method to pose

International Journal of Information Technology, Control and Automation (IJITCA) Vol.2, No.4, October 2012estimation. But, if the system's dynamic equation is non-linear and system noise is non-Gaussian, an approximation technique should be used. One of the most common numerical and approximation methods to pose estimation is sequential Monte Carlo [9]. This technique is known as particle filter [10] or Monte Carlo [11]. Let assume $X_t$ denotes the target's state at time $t$, and $Z_t$ denotes the set of all measurements $z_t$ up to time $t$. The main idea in particle filtering is representation of probability distribution function (pdf) by a set of weighted samples. Suppose that $S = \{(s^{(n)}, \omega^{(n)}) | n = 1,...,N\}$ is a set of weighted samples where $\sum_{n=1}^{N} \omega^{(n)} = 1$, then the posteriori density function is represented as follows.

$$p(X_t | Z_t) \approx \sum_{i=1}^{N} \omega_t^i \delta(X_t - X_t^i) \tag{1}$$

where $\omega_t^i$ is the weight of sample $i$ at time $t$ which is calculated by following equation.

$$\omega_t^i \propto \omega_{t-1}^i \frac{p(z_t | X_t^i) p(X_t^i | X_{t-1}^i)}{q(X_t^i | X_{t-1}^i, z_t)} \tag{2}$$

where $q(.)$ is importance density function which is used in sampling step. This method is called *Sampling Importance Sampling* (SIS). In the equation (1), $\delta$ is Kronecker delta function which is defined as follows:

$$\delta(t) = \begin{cases} 1 & t = 0 \\ 0 & t \neq 0 \end{cases} \tag{3}$$

In this paper, target's state is defined as follows:

$$X_t = \{(x_t^k, y_t^k) | k = 1,...,T\} \tag{4}$$

where $x_t^k$ and $y_t^k$ are coordinates of target $k$ at time $t$ in video screen, and $T$ is the number of targets. The particle filter method includes two main stages: *prediction* and *update*. In the prediction stage, particles of next step are obtained by using current particles and state transition equation. In the update stage, a weight is assigned to each particle using the available measurements. After the weight assignment, posteriori density function is calculated by equation (1). Then the target's next state is computed by a weighted average on the next step particles using following equation:

$$\hat{X}_t = \sum_{i=1}^{N} \omega_t^i X_t^i \tag{5}$$

39

## 3. SIR PARTICLE FILTER

One of the big challenges in implementation of SIS particle filter algorithm is appropriate choice of importance density function $q(.)$. This function is used in sampling step. Accuracy of pose estimation increases by increasing the similarity between importance density function and posteriori density function. Another problem in implementation of particle filter is degeneracy problem. After several iteration of particle filter algorithm, some of the particles have very small weight. This means that, some of the particles don't have effect on calculation of posteriori density function. This is an adverse event. There are two solutions to overcome the degeneracy problem: 1) appropriate choice of importance density function 2) resampling. Both of these solutions are used in SIR approach. State transition function is considered as importance density function.

$$q(X_t | X_{t-1}^i, z_t) = p(X_t | X_{t-1}^i) = N_{X_t}(X_t, \Sigma_t) \tag{6}$$

This is a suboptimal choice for importance density function. After state estimation in each step of SIR algorithm, current samples are resampled. After the resampling step, weights of samples are equal to $1/N$ ($N$ is the number of samples). If the state transition function is considered as importance density function, then the particle weights are calculated by following equation:

$$\omega_t^j \propto \omega_{t-1}^j p(z_t | X_t^i) \tag{7}$$

Equation (7) should be modified as follows, because weights of samples are equal to $1/N$ after the resampling step.

$$\omega_t^j = p(z_t | X_t^i) \tag{8}$$

where $p(z_t | X_t^i)$ is likelihood of each particle at time $t$.

## 4. TARGET REPRESENTATION MODEL

In this paper, each target's state is represented by a rectangle. Center of the rectangle denotes the target's current state, and dimensions of the rectangle are adjusted by target's size. Fig.1 shows the representation model of targets. In this figure, $H_x$ and $H_y$ are dimensions of surrounding rectangle of target object, and $(x, y)$ denotes the current state of each target. Model of each target consists of the target's state and dimensions of surrounding rectangle. If changes of the model become larger than the predetermined threshold then target's model will be updated. When the target model is updated, both target state and dimensions of corresponding rectangle will be updated.

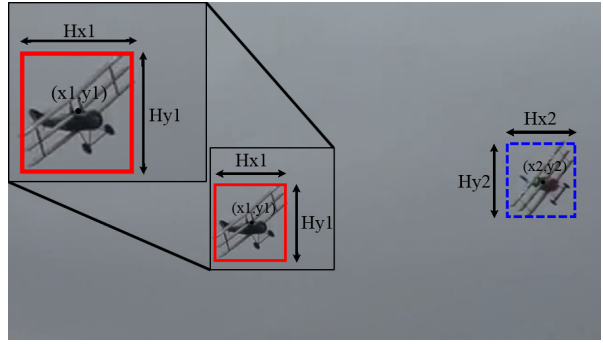

Figure 1. Target representation model.

## 5. SYSTEM DYNAMIC EQUATION

In this paper, we used real-world videos. These videos are extracted from real airshow videos that obtained from internet. In these videos, camera is movable. Sudden motion of camera causes undesirable motion in trajectory lines of targets. So, a discrete-time random walk process is considered as state transition equation for each target. In random walk model, the state transition equation is defined as follows.

$$X_t = X_{t-1} + v_t \qquad (9)$$

where $v_t \sim N(0, \Sigma_p)$ is a two dimensional Gaussian noise. Probability of state transition of each target is defined as following equation.

$$p(X_t | X_{t-1}) = N_{X_t}(X_t, \Sigma_t) \qquad (10)$$

## 6. LIKELIHOOD FUNCTION

We used combination of two functions to define the likelihood function in this study. First function calculates the likelihood according to difference between corresponding pixels in grayscale mode [16]. Second function calculates the likelihood of corresponding regions using a color histogram in RGB mode [13]. The likelihood function is defined as following equation:

$$p(z_t | X_t^i) = p_{intensity} + (p_{color})^2 \qquad (11)$$

In section 6.1, grayscale module is briefly explained, and an equation is defined to calculate $p_{intensity}$. In section 6.2, RGB module is introduced and an equation is defined to compute $p_{color}$.

### 6.1. Grayscale Module

Difference in intensity of corresponding pixels is used to calculate the similarity between target's region and each particle's region. If $S$ is the set of pixels inside the target's region and $C_j$ is the set of pixels inside the $j^{th}$ particle's region, then we calculate similarity of these regions using following equation [18]:

$$p_{intensity} = e^{\left(-\frac{\sum_{i=1}^{I}(T^i - C_j^i)}{H_x \times H_y}\right)} \qquad (12)$$

where $H_x$ and $H_y$ are the dimensions of rectangular region of target object, and $I$ is the number of pixels inside this region.

### 6.1. RGB Module

A color distribution model is used as measurement model in some studies [14]. This model is an 8×8×8 histogram in RGB color space. This histogram is calculated for target's rectangular region. Distance of each pixel from center of region affects values of histogram. In [14], the interval (0,255) is divided to 8 equal parts, and an 8×8×8 histogram is used. For reducing computational complexity, we used a 4×4×4 histogram. In this color distribution model, efficacy of each pixel on values of histogram reduces as it is placed far from the center of region. Epanechnikov kernel function is used to indicate this property in the histogram, which is defined as follows:

$$k(r) = \begin{cases} 1 - r^2 & r < 1 \\ 0 & otherwise \end{cases} \qquad (13)$$

where $r$ is the distance of each pixel from the center of region. The color distribution model is represented by a histogram which is shown in following equation:

$$p_s = \{p_s^{(u)}\}_{u=1,\ldots,m} \qquad (14)$$

where $s$ denotes the location of an arbitrary pixel, and $m$ is the number of histogram bins. Number of histogram bins is considered as $m=64$. The histogram value in location $s$ is obtained by following equation:

$$p_s^{(u)} = f \sum_{i=1}^{I} k\left(\frac{\|s - X_i\|}{a}\right) \delta(h(X_i) - u) \qquad (15)$$

where $I$ is the number of pixels inside the target's region, and $\|.\|$ is considered as Euclidean distance. $a = \sqrt{H_x^2 + H_y^2}$, where $H_x$ and $H_y$ are the dimensions surrounding rectangle of target object. In equation (15), $f$ is a normalization coefficient which is defined as follows:

$$f = \frac{1}{\sum_{i=1}^{I} k\left(\frac{\|s - X_i\|}{a}\right)} \tag{16}$$

Bhattacharyya distance is used to calculate the difference of two histograms using following equation:

$$d = \sqrt{1 - \rho[p,q]} \tag{17}$$

where $\rho[p,q]$ denotes the similarity of $p$ and $q$ and is defined as follows:

$$\rho[p,q] = \sum_{u=1}^{m} \sqrt{p^{(u)} \times q^{(u)}} \tag{18}$$

Finally, the distance between, histogram of target's region and histogram of each particle's region are used to assign the weight to each particle. $p_{color}$ is calculated by following equation:

$$p_{color} = e^{-\lambda d^2} \tag{19}$$

where λ is a parameter that experimentally considered *25* in this paper.

## 7. DATA ASSOCIATION

Data association algorithm associates the available measurements to detected targets. In cluttered environments, some of the obtained measurements may arise from clutter. A data association algorithm must detect the measurements that are arisen from clutter. Some of the important data association algorithms are Global Nearest Neighbor (GNN), Joint Probabilistic Data Association (JPDA) and Multi Hypothesis Tracking (MHT). GNN is more efficient algorithm for clutter-free environments. Our selected videos in this study had static background and our tracking environment was clutter-free. We used GNN as data association algorithm. Each measurement is associated to nearest track in GNN algorithm. Optimal association minimizes the sum of distances between each measurement and it's corresponding track. The optimal association can be obtained using a cost matrix and an optimization algorithm. Distance between each measurement and track represents their cost in this matrix. We used Munkres optimization algorithm in our study that was presented in [14].

This algorithm uses the cost matrix for determining the optimal association. We used three types of measurements: 1) grayscale module, 2) RGB module and 3) target detection for deformation detection. Grayscale module and RGB module don't require data association, because these are calculated for each particle separately. Data association is only used for target detection.

43

In our application, number of targets is constant and predetermined. If the number of detected targets in tracking environment is less than the predetermined number, then at least one of the targets has been occluded by another one. In such case, target detection step is ignored during the occlusion phenomenon. The color histogram and intensity of corresponding pixels are considered as observation. Data association won't be applied during the occlusion phenomenon, because data association is only used for target detection observation.

## 8. DEFORMATION DETECTION

Manoeuvring target tracking is one of the big challenges in target tracking. Target's model changes in occurrence of rotation or scaling. In such case, the target's old model is not efficient for weight assignment. Using static model for tracking manoeuvring targets don't yield good results. The MM approach is a commonly used approach to overcome this problem. In this paper, a dynamic model is used to improve the results of SIR algorithm in multiple manoeuvring target tracking. If changes of target's model are larger than the predetermined threshold, then the model should be updated. Old model of target is replaced by new one. Deformation detection is done using a color histogram that is described in section 6.2. After several frames, the considered targets are detected again by using a background subtraction algorithm. The color histogram is calculated for region of each detected target as described in section 6.2. Suppose that, the $q_{new}$ is the color histogram of new model of target and $q_{old}$ is the color histogram of old model of target. We used Bhattacharyya distance to detect the deformation of target model which is defined as following equation:

$$d = \sqrt{1 - \rho[q_{new}, q_{old}]} \tag{20}$$

where $\rho[q_{new}, q_{old}]$ denotes the similarity of two histograms and obtains using following equation:

$$\rho[q_{new}, q_{old}] = \sum_{u=1}^{m} \sqrt{q_{new}^{(u)} \times q_{old}^{(u)}} \tag{21}$$

when $d > T$, then changes of target's model is larger than a predetermined threshold and the its previous model should be replaced by new one. $T$ is a threshold which is experimentally considered as *0.12*. Our proposed DDPF algorithm is described in Table 1.

## 9. CONDITIONS OF EXPERIMENTS

Our proposed DDPF algorithm and basic SIR algorithm are implemented by MATLAB 2011(b). These algorithms are tested on a system with configuration: CORE i7 CPU, 4GB RAM and 1GB Geforce VGA. The real-world videos are selected to test our proposed algorithm. These videos are extracted from parts of real airshow videos. Format of these videos is "mp4", and their dimensions are 240×320. It is assumed that the number of targets is constant and predetermined in these videos. A background subtraction algorithm is used to detect the targets, because the selected videos have partly static background. Background of selected videos should be reclusion. Our proposed DDPF method is appropriate for tracking in clutter-free environment.

## 10. COMPARISON

In this section, we compare results of our proposed DDPF algorithm with basic SIR particle filter algorithm in multi manoeuvring target tracking. Fig.2 shows that the basic SIR-PF algorithm is not able to track the manoeuvring targets without using dynamic model for targets. Fig.3 shows that DDPF algorithm efficiently tracks the manoeuvring targets. Comparison results show that the basic SIR-PF algorithm misses the targets when the rotation or scaling occurs in target's model, whereas proposed DDPF method resolves this problem by updating the model when the rotation or scaling occurs. Results of our proposed method are very sensitive to parameters settings. By increasing the dimensions of videos, the computational complexity of our approach increases drastically. We tested our proposed approach on other videos and their results are shown in Fig.4 and Fig.5.

Table 1. Deformation Detection Particle Filter algorithm

Detect the desired target using background subtraction and calculate the reference color histogram $q^*$ for target's model by following equation and generate N particles $S_1 = \{X_1^i | i = 1, ..., N\}$.

$$q^* = f \sum_{j=1}^{J} k\left(\frac{\|s - X_j\|}{a}\right) \delta(h(X_j) - u)$$

FOR frame = 1:Number
1) IF frame % 5 = 0
   Run deformation detection algorithm for current frame and if d>T then replace old model by new model and re-calculate $q^*$ for new model.
   END IF
2) **Prediction:** Generate N new particles $S_{t+1} = \{X_{t+1}^i | i = 1, ..., N\}$ from previous particle set $S_t = \{X_t^i | i = 1, ..., N\}$ using dynamic system model.
3) Compute the $p_{color}$ and $p_{intensity}$ for each particle by equations (12) and (19).
4) **Weighting:** Calculate the weight for each particle using following equation.

$$\omega_{t+1}^i = p_{intensity}^i + (p_{color}^i)^2$$

5) Normalize weights as follows:

$$\hat{\omega}_{t+1}^i = \frac{\omega_{t+1}^i}{\sum_{j=1}^{N} \omega_{t+1}^j}$$

6) Calculate the estimated state using weighted average on the particle set $S_{t+1} = \{(X_{t+1}^i, \omega_{t+1}^i) | i = 1, ..., N\}$.

$$\hat{X}_{t+1} = \sum_{i=1}^{N} \omega_{t+1}^i X_{t+1}^i$$

7) **Resampling:** Generate N particle $\hat{S}_{t+1} = \{(X_{t+1}^i, \frac{1}{N}) | i = 1, ..., N\}$ from the $S_{t+1} = \{(X_{t+1}^i, \omega_{t+1}^i) | i = 1, ..., N\}$ by CSW approach [12].
END FOR

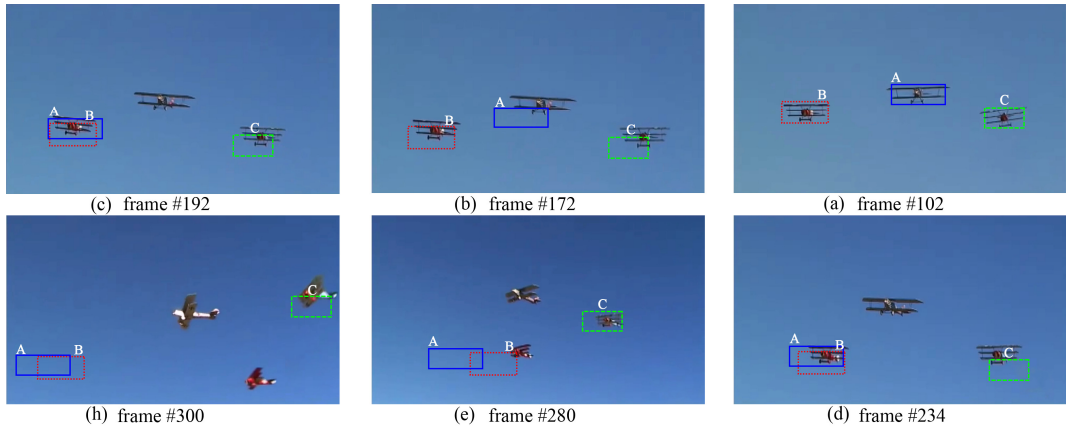

(c) frame #192    (b) frame #172    (a) frame #102

(h) frame #300    (e) frame #280    (d) frame #234

Figure 2. Result of basic SIR algorithm in multiple manoeuvring targets tracking without using dynamic target's model.

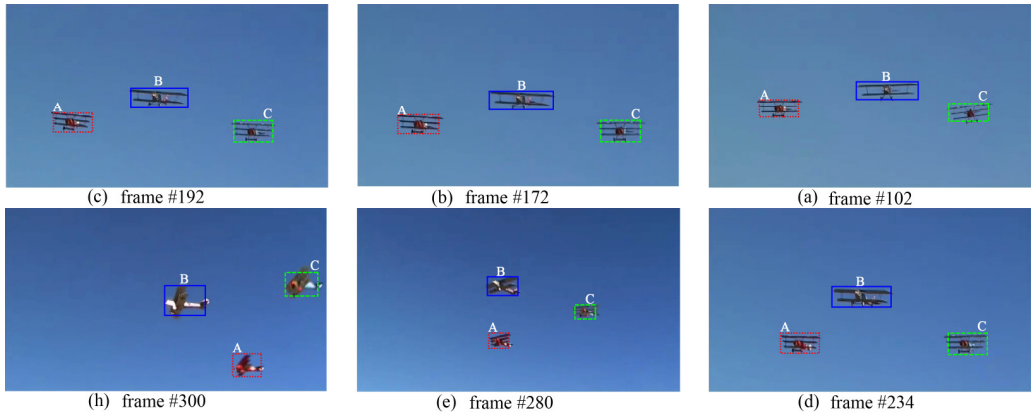

Figure 3. Result of DDPF method in multiple manoeuvring targets tracking using dynamic target's model.

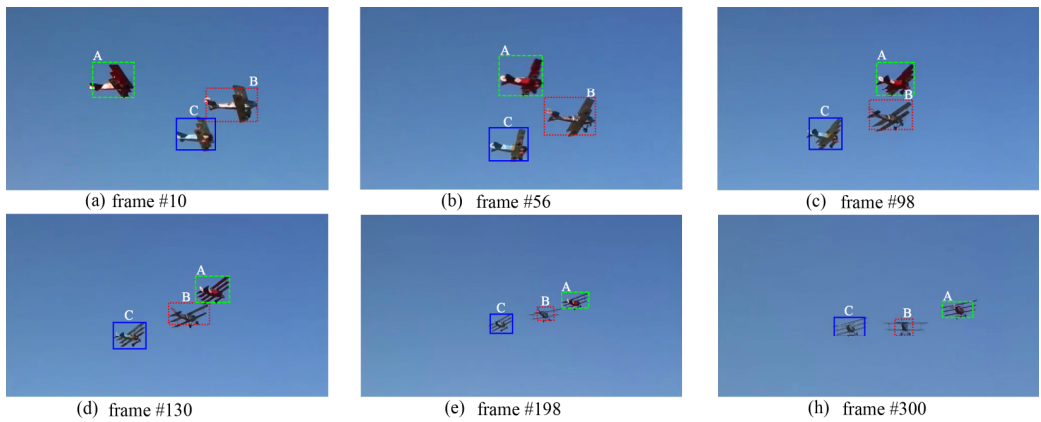

Figure 4. Result of DDPF method in multiple manoeuvring targets tracking without occlusion.

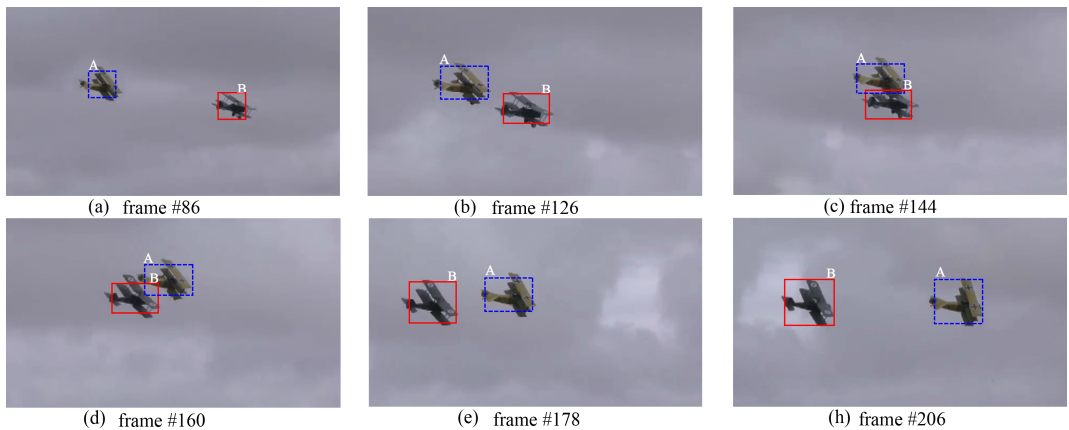

Figure 5. Result of DDPF method in tracking two manoeuvring aircrafts with partial occlusion.

## 11. CONCLUSION AND DISCUSSION

In this paper, we proposed DDPF method by applying a dynamic model for targets in the SIR based particle filtering algorithm for tracking multiple manoeuvring targets. Results illustrate that the basic SIR-PF algorithm is not able to efficiently tracks the multiple manoeuvring targets when the rotation or scaling is occurred in target model, whereas DDPF method will be able to track multiple manoeuvring targets more robustly. Our studies showed that the results are very sensitive to parameters settings. Extending this approach by adaptive parameters is under study as our future work. One of the problems in multi manoeuvring target tracking is occlusion problem. Resolving the occlusion problem for improving the results of proposed approach is our other future work.

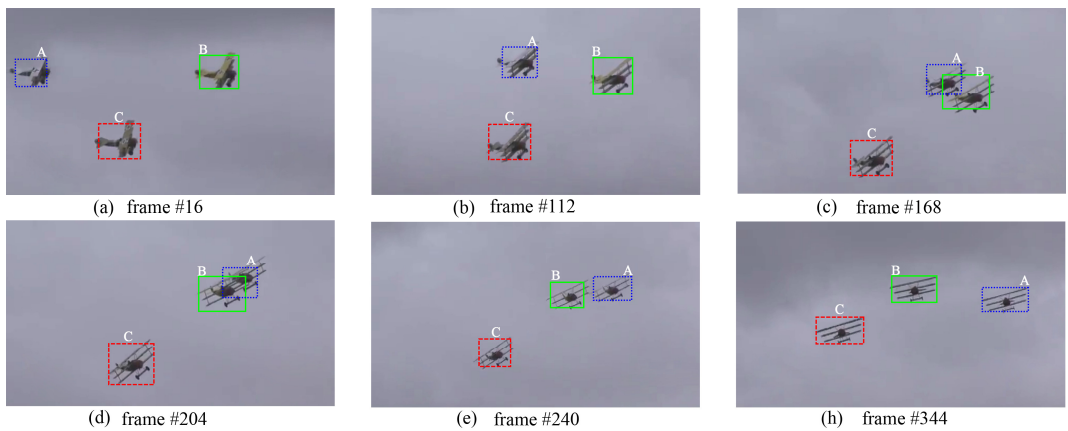

Figure 6. Result of DDPF method in tracking three manoeuvring aircrafts with partial occlusion.

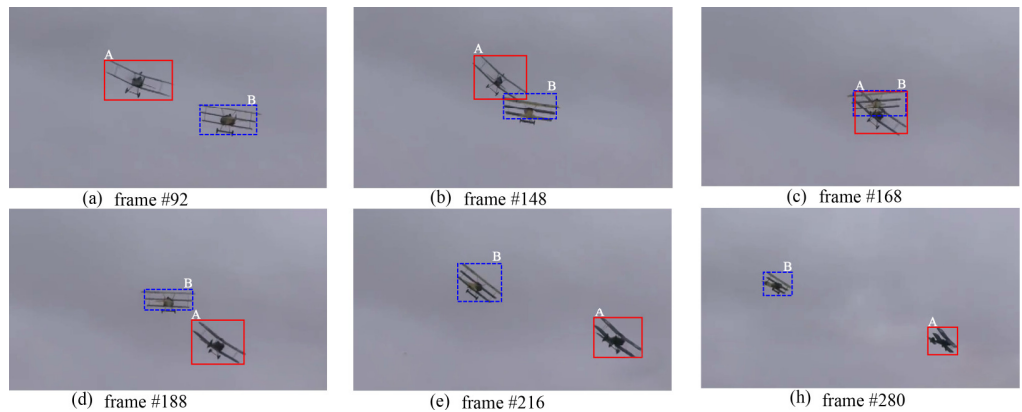

Figure 7. Result of DDPF method in tracking two manoeuvring aircrafts with complete occlusion.

**Authors**

**Mohammad Javad Parseh** is M.Sc. student of Artificial Intelligence in University of Tabriz in Iran. He received his B.Sc. in Software Engineering from Chamran university of Ahwaz in 2008. His research interests are Bayesian estimation, target tracking and state estimation, ontology matching in semantic web, fuzzy decision making and operating system.

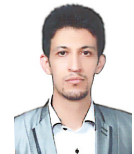

**Saeid Pashazadeh** is Assistant Professor of Software Engineering and chair of Information Technology Department at Faculty of Electrical and Computer Engineering in University of Tabriz in Iran. He received his B.Sc. in Computer Engineering from Sharif Technical University of Iran in 1995. He obtained M.Sc. and Ph.D. in Computer Engineering from Iran University of Science and Technology in 1998 and 2010 respectively. He was Lecturer in Faculty of Electrical Engineering in Sahand University of Technology in Iran from 1999 until 2004. His main interests are development, modelling and formal verification of distributed systems, computer security, particle filtering and wireless sensor/actor networks. He is member of IEEE and senior member of IACSIT and Member of editorial board of journal of electrical engineering of University of Tabriz in Iran.

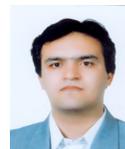